\documentclass[nohyperref]{article}

\usepackage{microtype}
\usepackage{graphicx}
\usepackage{subfigure}
\usepackage{booktabs} 

\usepackage{hyperref}

\usepackage[accepted]{icml2022}

\usepackage{amsmath}
\usepackage{amssymb}
\usepackage{mathtools}
\usepackage{amsthm}

\usepackage{relsize}

\usepackage{lipsum} 

\usepackage{hyperref}

\usepackage[capitalize,noabbrev]{cleveref}

\theoremstyle{plain}

\theoremstyle{definition}

\theoremstyle{remark}

\usepackage[textsize=tiny]{todonotes}

\icmltitlerunning{DiPietro-Hazari Kappa}

\begin{document}
\onecolumn
\icmltitle{DiPietro-Hazari Kappa: A Novel Metric for Assessing Labeling\\Quality via Annotation}



\icmlsetsymbol{equal}{*}

\begin{icmlauthorlist}
\icmlauthor{Daniel M. DiPietro*}{}
\icmlauthor{Vivek Hazari*}{}
\end{icmlauthorlist}


\icmlcorrespondingauthor{Daniel DiPietro}{dipietrodaniel131@gmail.com}

\icmlkeywords{Machine Learning, Computational Physics}

\vskip 0.3in



\printAffiliationsAndNotice{\icmlEqualContribution} 

\begin{abstract}

Data is a key component of modern machine learning, but statistics for assessing data label quality remain sparse in literature. Here, we introduce DiPietro-Hazari Kappa, a novel statistical metric for assessing the quality of suggested dataset labels in the context of human annotation. Rooted in the classical Fleiss's Kappa measure of inter-annotator agreement, the DiPietro-Hazari Kappa quantifies the the empirical annotator agreement differential that was attained above random chance. We offer a thorough theoretical examination of Fleiss's Kappa before turning to our derivation of DiPietro-Hazari Kappa. Finally, we conclude with a matrix formulation and set of procedural instructions for easy computational implementation\footnote{The DiPietro-Hazari Kappa is available as a high-performance Python function at \href{https://github.com/dandip/DH\_Kappa}{https://github.com/dandip/DH\_Kappa}}.

\end{abstract}

\section{Introduction}

While the datasets powering state-of-the-art machine learning models have grown exponentially in size, the metrics used to assess the quality of these datasets have remained stagnant and unnuanced. Here, we present a novel statistical metric capable of assessing dataset quality for supervised learning tasks.

Datasets used in supervised learning tasks consist of a target label and a vector of features. While the target label is generally treated as a ground truth, it is often created by unwritten heuristics used when assembling the dataset, such as the source of the data or keywords used to obtain it. These heuristics are imperfect, and it is important to quantitatively assess their quality. One method for doing this is to randomly select a subset of the data and have multiple human annotators label it. Generally, these annotations are then assessed via measures of \textit{inter-annotator agreement}. If annotators are generally in strong agreement with each other over the aggregate of the dataset, then the inter-annotator agreement is high. If they are not, then the inter-annotator agreement is low. Datasets with high inter-annotator agreement are assumed to be of high quality, while those without are assumed to be of low quality.

Unfortunately, current methods of inter-annotator agreement do not taken into account the suggested label for each piece of data. In other words, annotators can entirely disagree with heuristic-suggested labels, but, so long as they agree with each other, the inter-annotator agreement is high. However, clearly the suggested label would not be of high quality if all of the annotators (uniformly) disagreed with it. So, traditional measures of inter-annotator agreement have little value in this context.

It's worth noting that this suggested label need not come from some simple dataset generation heuristic. Indeed, a statistical metric for assessing labels in the context of human annotation could also be used to assess the performance of \textit{any} proposed label, including those produced from machine learning classification models on unlabeled, novel data. Note that this assumes that human annotators perform well on the task being assessed, as it is pointless to assess label quality using human annotators if the annotators themselves are incapable.

The requirements for an inter-annotator agreement metric assessing pre-labeled data are simple:
\begin{enumerate}
    \item if annotators agree on the suggested label (and, by extension, with each other), then the suggested label is good.
    \item if annotators disagree with the suggested label but do not agree with each other on their label of choice, the suggested label is poor.
    \item if annotators disagree with the suggested label and agree with each other on their label of choice, the suggested label is very poor.
\end{enumerate}

This paper proposes the \textit{DiPietro-Hazari Kappa} ($\kappa_{DH}$), a novel statistical measure that assesses the quality of suggested dataset labels in the context of inter-annotator agreement, quantitatively implementing the three requirements outlined above. We begin with an intuitive presentation of \textit{Fleiss's Kappa}, a common measure of inter-annotator agreement that serves as the foundation for our derivation. We then turn to a theoretical presentation of our novel metric.

\section{Fleiss's Kappa: Building an Intuition}

The DiPietro-Hazari Kappa is heavily rooted in the intuition of the Fleiss's Kappa ($\kappa$) metric \cite{fleiss1971measuring}. Thus, we offer an extensive presentation of Fleiss's Kappa below.

Suppose we have $n$ pieces of data, each denoted $d_1, \dots, d_n$. Each piece of data is assessed by $N$ annotators. There are $m$ possible categories, each denoted $c_1, \dots, c_m$. 

Consider the function $P(d_i, c_j)$. Given a data point $d_i$ and a category $c_j$, this function yields the number of annotators that placed $d_i$ in category $c_j$.

\subsection{Expected Pair Agreement}

The first step of computing Fleiss's Kappa is to find the proportion of total annotations that each category accounts for. In other words, what is the likelihood that a human annotator will place a piece of data in each category if they label in accordance with the distribution of annotations over the entire dataset? We denote this value for each category as $C_1, \dots, C_m$. To compute these values, we use
\begin{equation}
    \textbf{$C_j = \frac{1}{Nn} \sum_{i=0}^n P(d_i, c_j)$}    
\end{equation}

Now, based on this distribution, what is the chance that two annotators agree on any category by sheer randomness? Well, there is a $C_j$ chance of a single annotator predicting category $j$. So, there is a $C_j \cdot C_j=C_j^2$ chance of two annotators both predicting category $j$, assuming they label independently and in accordance with the distribution of labels over the dataset. By extension, there is a $C_E=\sum_{j=0}^m C_j^2$ chance that two annotators agree on any category based on our distribution. So, $C_E$ serves as a useful baseline for measuring inter-annotator agreement: agreement is only meaningful if it is happening at a rate above what we would expect from random chance.

\subsection{Observed Pair Agreement}

The second step of computing Fleiss's Kappa is to find the proportion of total possible annotator pairs that actually agreed for each data point. We denote these values as $R_1, \dots, R_n.$ These values are computed by dividing the number of annotator pairs that empirically agreed by the total number of possible pairs. For each data point, there are ${N \choose 2}$ possible annotator pairs. For a data point $d_i$, there are $\sum_{j=0}^m {P(d_i, c_j) \choose 2}$ pairs that actually agreed. Then, we define this proportion as follows
\begin{equation}
    R_i = \frac{1}{{N \choose 2}} \sum_{j=0}^m {P(d_i, c_j) \choose 2}.
\end{equation}

Now, we compute $\overline{R}$, which is the the average of $R_1, \dots, R_n$. In other words, $\overline{R}$ is the observed rate of annotator agreement across the entire dataset.

\subsection{Agreement Observed above Random Chance}

Now, we have two key values. We have $C_E$, which indicates the rate of annotator agreement that is expected by sheer chance given the annotation distribution over the entire dataset. We also have $\overline{R}$, which represents the observed rate of annotator agreement across our dataset.

Consider the value $1-C_E$. $1$ indicates perfect annotator agreement--$100\%$ of annotators agreed with each other. Thus, $1-C_E$ indicates \textit{the maximum attainable agreement above random chance}.

Now, consider the value $\overline{R}-C_E$. This demonstrates the agreement that was obtained \textit{in practice} above sheer randomness.

We may now compute Fleiss's Kappa as
\begin{equation}
    \kappa=\frac{\overline{R}-C_E}{1-C_E}
\end{equation}
Again, the denominator indicates the agreement that is achievable above random chance. The numerator indicates the agreement that was achieved in practice above random chance. As a result, Fleiss's Kappa is the proportion of possible performance above random chance that was achieved. Thus, a Fleiss's Kappa of $1$ indicates that we achieved perfect agreement, whereas a negative Fleiss's Kappa indicates that our inter-annotator agreement underperformed what would be expected by chance.

\section{DiPietro-Hazari Kappa: Derivation and Intuition}

The DiPietro-Hazari Kappa calculation proceeds similarly as above, except with the addition of \textit{suggested labels} that we would like to assess.

Suppose we have $n$ pieces of data, each denoted $d_1, \dots, d_n$. Each piece of data is assessed by $N$ annotators. There are $m$ possible categories, each denoted $c_1, \dots, c_m$. We also have $n$ ``proposed labels'', each denoted $\ell_1, \dots, \ell_n$ where each $\ell_i \in \{c_j\}_{j=1}^m$ denotes the proposed category label of $d_i$. Define the function $P(d_i, c_j)$ as above. Define the function $Q(\ell_i, c_j) = 1$ if $\ell_i = c_j$ and $0$ otherwise.

\subsection{Expected Correct Pair Agreement}

First, compute the proportion of total annotations that each category accounts for, as is done for Fleiss's Kappa. We denote this value for each category as $C_1, \dots, C_m$. To compute these values, we use
\begin{equation}
    \textbf{$C_j = \frac{1}{Nn} \sum_{i=0}^n P(d_i, c_j)$}    
\end{equation}

Next, compute the proportion of proposed labels that each category accounts for, denoted $L_1, \dots, L_m$, as follows
\begin{equation}
    L_j = \frac{1}{n} \sum_{i=0}^n Q(\ell_i, c_j)
\end{equation}

Now, $L_j$ gives us the chance that a randomly selected point was given a proposed label of category $c_j$ if labeled at random given the distribution of suggested labels over the dataset. $C_j$ gives us the chance that an annotator randomly guesses category $c_j$. Then, the chance that an annotator randomly guesses a category $c_j$ and is correct is $C_j \cdot L_j$. The chance that two annotators guess category $c_j$ and are correct (and also agree with each other) is $C_j^2 \cdot L_j$. Then, the chance that two annotators agree on the correct label \textit{for any category} by sheer chance is
\begin{equation}
    C_E = \sum_{j=1}^m C_j^2 \cdot L_j.
\end{equation}

\subsection{Expected Incorrect Pair Agreement}

We would like to find the chance that two annotators disagree with the proposed label but agree with each other \textit{by chance}. 

First, we have
\begin{equation}
    \sum_{j=1, j\ne {j'}}^{m} C_j^2    
\end{equation}
which yields the chance that two annotators agree at random on a category that is \textit{not} $c_{j'}$. Recall that $L_{j'}$ gives the chance that a randomly selected point was given a proposed label category of $c_{j'}$. So, the probability that a randomly selected point is given a proposed label of $c_{j'}$ and two annotators agree on any category that isn't $c_{j'}$ by sheer chance is
\begin{equation}
    L_{j'}\sum_{j=1, j\ne {j'}}^{m} C_j^2.
\end{equation}
Then, the chance that two annotators agree on a label other than the proposed label \textit{for any category} by sheer chance is $C_F$, defined as follows
\begin{equation}
    C_F = \sum_{i=1}^m\left( L_i\sum_{j=1, j\ne i}^{m} C_j^2 \right)    
\end{equation}

\subsection{Observed Correct Pair Agreement}

Now, we would like to find the proportion of annotator pairs that agreed with each other and the proposed label for each data point. We denote this value $R_1, \dots, R_n$. This value is computed by dividing the number of pairs that agreed on the correct label by the number of total possible pairs.

For each data point, there are ${N \choose 2}$ possible annotator pairs. For a data point $d_i$, there are ${P(d_i, \ell_i) \choose 2}$ pairs that actually agreed on the correct label. Then, we have
\begin{equation}
    R_i = \frac{1}{{N \choose 2}} {P(d_i, \ell_i) \choose 2}    
\end{equation}

Now, we compute $\overline{R}$, which indicates the average of $R_1, \dots, R_n$. In other words, $\overline{R_E}$ is the rate of annotator agreement on the proposed label across the entire dataset.

\subsection{Observed Incorrect Pair Agreement}

Next, we would like to find the proportion of annotator pairs that agreed with each other but disagreed with the proposed label. We denote this value $S_1, \dots, S_n$.

For each data point, there are ${N \choose 2}$ possible annotator pairs. For a data point $d_i$, there are 
\begin{equation}
    \sum_{j=1, c_j \ne \ell _i}^n {P(d_i,c_j) \choose 2}
\end{equation}
pairs that agreed on a specific label other than the proposed label. Then, we have
\begin{equation}
    S_i = \frac{1}{{N \choose 2}} \left( \sum_{j=1, c_j \ne \ell _i}^n {P(d_i,c_j) \choose 2} \right) 
\end{equation}

Now, we compute $\overline{S}$, which indicates the average of $S_1, \dots, S_n$. In other words, $\overline{S}$ is the rate of annotator agreement on a label other than the proposed label, measured across the entire dataset.

\subsection{Agreement Differential Obtained above Chance}

Recall that $C_E$ indicates the proportion of annotators that we expect to agree with each other and the proposed label by sheer chance. Similarly, recall that $C_F$ indicates the proportion of annotators that we expect to agree with each other but disagree with the proposed label by sheer chance. Then, we refer to $C_E - C_F$ as the \textit{expected annotator agreement differential}.

Consider the value $1-(C_E-C_F)$. $1$ indicates perfect annotator agreement on the correct label and no annotator agreement on the incorrect label. Thus, $1-(C_E-C_F)$ indicates \textit{the maximum attainable agreement differential above random chance}.

Note that $\overline{R}-\overline{S}$ indicates the annotator agreement differential observed in practice. Then, $(\overline{R}-\overline{S}) - (C_E-C_F)$ indicates the agreement differential that was obtained \textit{in practice} above sheer randomness. Hence, we construct the DiPietro-Hazari Kappa so that it indicates the proportion of possible agreement differential above chance that was achieved in practice.

\begin{equation}
    \kappa_{DH} = \frac{(\overline{R}-\overline{S})-(C_E-C_F)}{1-(C_E-C_F)}    
\end{equation}

\section{DiPietro-Hazari Kappa: Matrix Formulation for Easy Computation}

In this section, we offer a concise matrix formulation of the DiPietro-Hazari Kappa to enable convenient computational implementation.

Let there be $n$ pieces of data $d_1, \dots, d_n$ with $m$ possible categories, $c_1, \dots, c_m$. Each piece of data is assessed by $N$ annotators; there are $n$ proposed labels as well, denoted $\ell_1, \dots, \ell_n$, where each $\ell_i \in \{c_j\}_{j=1}^m$ denotes the proposed category label of $d_i$. Let $A$ define the $n \times m$ matrix where each $a_{ij}$ indicates the number of annotators that placed $d_i$ in $c_j$. Let $B$ denote the $n \times m$ matrix where $b_{ij}=1$ if $\ell_i = c_j$ and $0$ otherwise. Define $\oplus$ as the Hadamard product. Let $f(P\mapsto Q)$ define the element-wise matrix map $p_{ij} \mapsto {p_{ij} \choose 2}$ where $p_{ij} \in P$, ${p_{ij} \choose 2} = q_{ij} \in Q$. Let $\text{row\_sum}(P)$ and $\text{col\_sum}(P)$ denote row-wise and column-wise matrix summation functions respectively. Now, we may compute $\kappa_{DH}$ via the procedural instructions in (14).

\begin{equation}
\begin{split}
    C & \leftarrow (\text{col\_sum}(A)) \cdot \frac{1}{Nn} \\
    L & \leftarrow (\text{col\_sum}(B)) \cdot \frac{1}{n} \\
    C_E & \leftarrow \text{row\_sum}(C \oplus C \oplus L) \\
    T & \leftarrow \begin{bmatrix}
    C \\
    \vdots\\
    C\\
    \end{bmatrix}_{m \times m} \oplus \left( 
    \mathbf{J}_{m \times m} - \mathbf{I}_{m \times m}
    \right)\\
    C_F & \leftarrow \text{col\_sum} ( \text{row\_sum}(T \oplus T) \oplus L^T )\\
    \overline{R} & \leftarrow \text{col\_sum} \left( \text{row\_sum} ( f(A) \oplus B ) \cdot \frac{1}{{N \choose 2}} \right) \cdot \frac{1}{n}\\
    \overline{S} & \leftarrow \text{col\_sum} \left( \text{row\_sum} ( f(A) \oplus (\mathbf{J}_{n \times m} - B) ) \cdot \frac{1}{{N \choose 2}} \right) \cdot \frac{1}{n}\\
    \kappa_{DH} & \leftarrow \frac{(\overline{R}-\overline{S})-(C_E-C_F)}{1-(C_E-C_F)}    
\end{split}
\end{equation}

These instructions are implemented as a high-efficiency Python function, available at \href{https://github.com/dandip/DH\_Kappa}{https://github.com/dandip/DH\_Kappa}.

\section{Conclusion}

Here, we presented thorough theoretical outlines of Fleiss's Kappa, as well as our novel metric DiPietro-Hazari Kappa. To our knowledge, this is the first statistical measure that uses inter-annotator agreement to assess the quality of a dataset's labels (which may be generated by a heuristic or inferenced by a model). As the importance of dataset quality becomes increasingly apparent in machine learning research, this metric has the potential to serve as a commonplace benchmark in supervised learning literature.

\section*{Acknowledgments}

We thank Ziray Hao, John McCambridge, and Alexander ``Sasha'' Kokoshinskiy for valuable discussions concerning the creation of this metric.

\bibliography{bib}
\bibliographystyle{icml2022}


\end{document}